\documentclass[conference,compsoc]{IEEEtran}
\usepackage{graphicx}
\usepackage{amssymb}
\usepackage{amsmath}
\usepackage{subfigure}
\usepackage{multirow}
\usepackage{booktabs}
\usepackage{multicol}
\usepackage[linesnumbered,boxed,ruled,commentsnumbered]{algorithm2e}

\ifCLASSOPTIONcompsoc
  \usepackage[nocompress]{cite}
\else
  \usepackage{cite}
\fi

\ifCLASSINFOpdf
\else
\fi

\begin{document}
\title{Temporal Reasoning Graph for Activity Recognition}

\author{\IEEEauthorblockN{Jingran~Zhang}
\IEEEauthorblockA{Center for Future Multimedia\\
School of Computer Science and Engineering\\
University of Electronic Science and Technology of China\\
Chengdu 610051, China\\
jrzhang339@gmail.com}
\and
\IEEEauthorblockN{Fumin~Shen}
\IEEEauthorblockA{Center for Future Multimedia\\
School of Computer Science and Engineering\\
University of Electronic Science and Technology of China\\
Chengdu 610051, China\\
fumin.shen@gmail.com}
\and
\IEEEauthorblockN{Xing~Xu}
\IEEEauthorblockA{Center for Future Multimedia\\
School of Computer Science and Engineering\\
University of Electronic Science and Technology of China\\
Chengdu 610051, China\\
xing.xu@uestc.edu.cn}
\and
\IEEEauthorblockN{Heng~Tao~Shen}
\IEEEauthorblockA{Center for Future Multimedia\\
School of Computer Science and Engineering\\
University of Electronic Science and Technology of China\\
Chengdu 610051, China\\
shenhengtao@hotmail.com}
}


\maketitle

\begin{abstract}
Despite great success has been achieved in activity analysis, it still has many challenges. Most existing work in activity recognition pay more attention to design efficient architecture or video sampling strategy. However, due to the property of fine-grained action and long term structure in video, activity recognition is expected to reason temporal relation between video sequences. In this paper, we propose an efficient temporal reasoning graph (TRG) to simultaneously capture the appearance features and temporal relation between video sequences at multiple time scales. Specifically, we construct learnable temporal relation graphs to explore temporal relation on the multi-scale range. Additionally, to facilitate multi-scale temporal relation extraction, we design a multi-head temporal adjacent matrix to represent multi-kinds of temporal relations. Eventually, a multi-head temporal relation aggregator is proposed to extract the semantic meaning of those features convolving through the graphs. Extensive experiments are performed on widely-used large-scale datasets, such as Something-Something and Charades, and the results show that our model can achieve state-of-the-art performance. Further analysis shows that temporal relation reasoning with our TRG can extract discriminative features for activity recognition.
\end{abstract}


% \fi
\IEEEpeerreviewmaketitle

\section{Introduction}

    \begin{figure}[t]
        \centering
        \includegraphics[width=0.99\columnwidth]{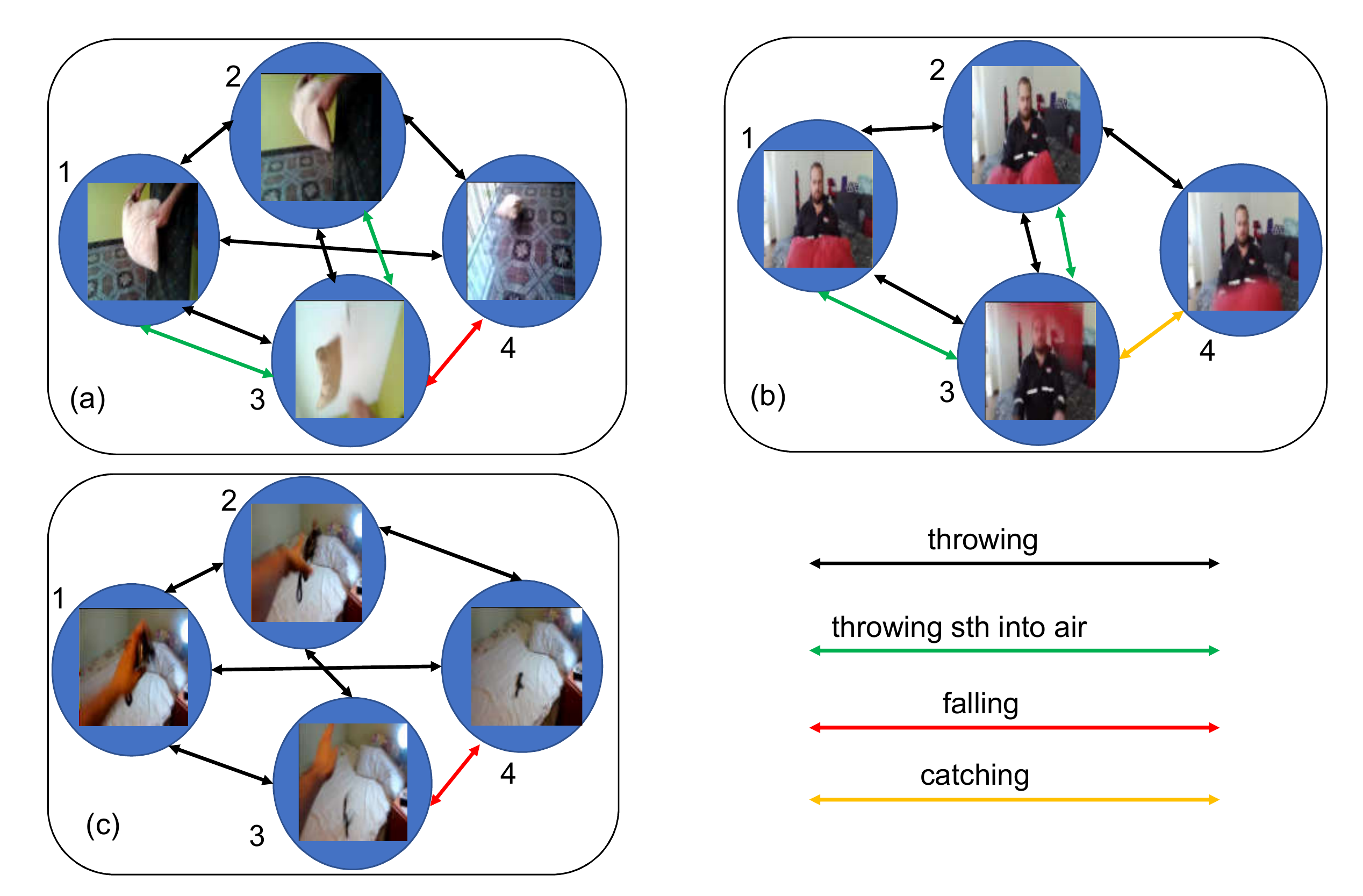}
        \caption{The example of temporal relation helps to classify and distinguish an activity (all the samples are originated from Something-Something V2 dataset). The three kinds of activities are (a) throwing something into the air and letting it fall; (b) throwing something into the air and catching it; (c) throwing something. We cannot distinguish one activity from the other two with a single frame. The inference of the temporal relation between the sequence is needed for deducing what has happened and classifying an activity.}
        \label{fig:example}
    \end{figure}
    
    As one of the important video analysis tasks, activity recognition has attracted significant attention from the academic community in computer vision. Thanks to the efficiency of deep learning techniques and larger datasets in computer vision, the action recognition performance has been remarkably boosted. Two kinds of architectures are widely adopted: (1) 2 dimensional convolution neural networks (2D ConvNets, 2D CNNs) for capturing frame-level features \cite{wang2016temporal}; (2) 3 dimensional convolution neural network (3D ConvNets, 3D CNNs) \cite{carreira2017quo, qiu2017learning} or recurrent neural networks (RNNs) \cite{donahue2015long} for modeling temporal context information. Afterward, a natural idea to boost the recognition performance based on the above architectures is to applying two-stream \cite{simonyan2014two} based framework which has heterogeneous inputs for action recognition. However, those approaches cannot capture the underlying structure which is characterized by transformations and temporal relations rather than the appearance of certain entities \cite{zhou2018temporal}. Therefore, they are very inefficient for activity recognition, because the activity can be characterized by the temporal evolution of appearance governed by motion. Consider three kinds of frequently occurred activity in ordinary daily life as shown in Figure \ref{fig:example}. Those activities cannot be recognized without reasoning about short or long-term temporal relations. An ordinary activity typically consists of several temporal relations at a multi-scale time-span. As the example shown in Figure \ref{fig:example} (a) and (b), the activity ``throwing'' contains the short-term relation like ``pillow throwing'' and ``falling'', and long-term relation like ``keeping it falling'' or ``catching it'', i.e., activity often equipped with specific spatial patterns as well as multi-scale temporal structure.
    
    % \begin{figure}[t]
    %     \centering
    %     \includegraphics[width=0.99\columnwidth]{example.pdf}
    %     \caption{The example of temporal relation helps to classify and distinguish an activity (all the samples are originated from Something-Something V2 dataset). The three kinds of activities are (a) throwing something into the air and letting it fall; (b) throwing something into the air and catching it; (c) throwing something. We cannot distinguish one activity from the other two with a single frame. The inference of the temporal relation between the sequence is needed for deducing what has happened and classifying an activity.}
    %     \label{fig:example}
    % \end{figure}
    
    \begin{figure*}[h]
    \centering
    \includegraphics[width=0.99\textwidth]{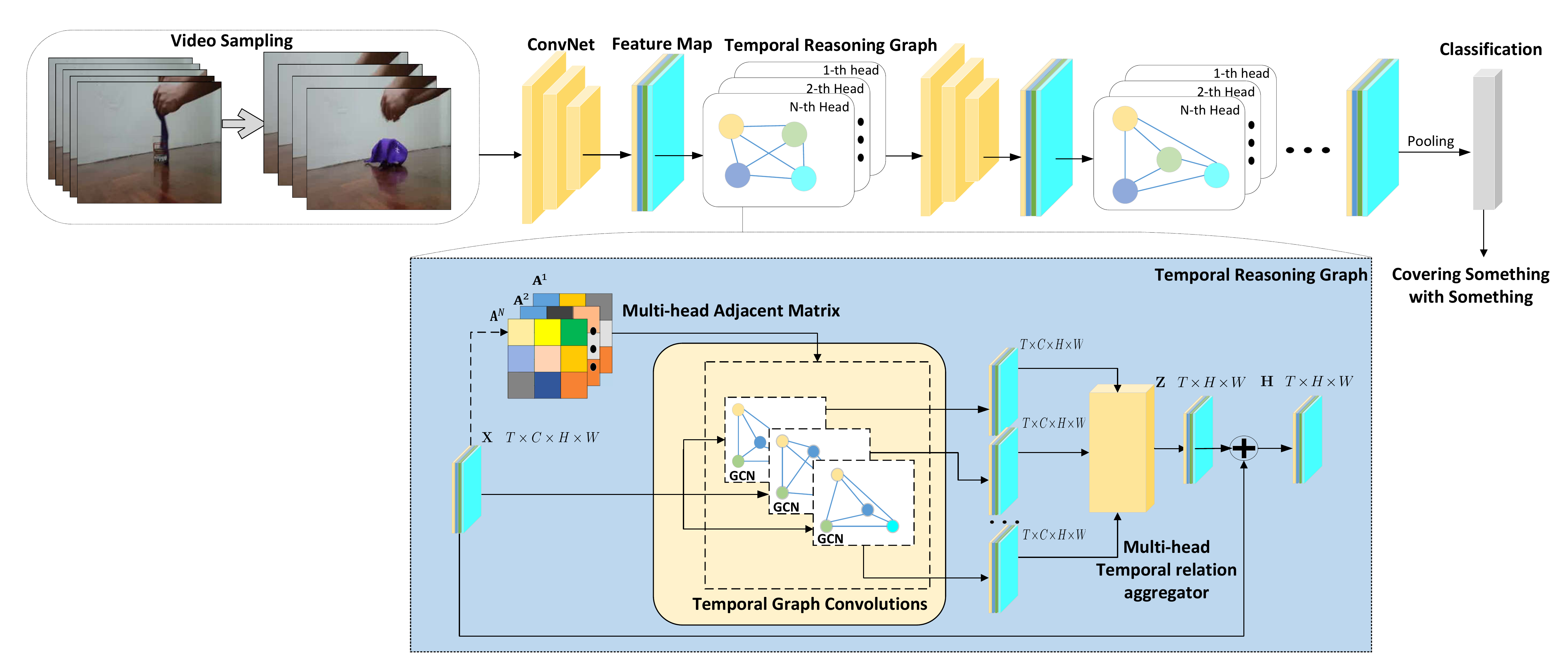}
    \caption{The overall architecture of our proposed temporal graph reasoning network. We apply ConvNets to extract video features. Afterward, we perform graph convolution on the temporal sequence with the learnable multi-head adjacent matrix and aggregate the semantic temporal features with the temporal relation aggregator. Finally, the spatial and temporal features are concatenated into a compact form for the final activity.}
    \label{fig:model}
    \end{figure*}
    
    Hence, the ability to accurately capture relevant relation between temporal sequence and perform temporal reasoning is crucial for understanding activity in video. It is expected to discover temporal semantic knowledge of a video beyond the appearance of objects in the frames over short and long-range temporal dependencies. Therefore, to effectively exploit such relation, it is required to develop an effective deep model that has the capacity to cope with temporal semantic meaning for activity understanding. Several works \cite{wang2018appearance-and-relation, zhou2018temporal} have attempted to explore the relation between video sequences. They apply multi-layer perceptron (MLP) which may have limited capacity for learning similarity of sequence object to investigate the relation between video sequence object. Additionally, they could only explore short-range temporal dependencies relation and cannot capture the transformable temporal states over time.
    
    In this paper, we propose a \textit{temporal reasoning graph (TRG)} network by directly performing temporal state relation reasoning in the graph to address the problem of capturing the transformable temporal states and different time-scale action dependencies. Notably, graph convolution networks (GCNs) \cite{stgcn2018aaai, zhou2018graph} are impressively powerful tools to analyze relations, but they typically have identified adjacent relation of different objects. Our proposed TRG approach tackles the key problem of how to represent the temporal graph connection structure. Inspired by the temporal relation network \cite{zhou2018temporal} which defines pairwise temporal relation as a composite function and exploits MLP to capture their semantic dependency, in this paper, the proposed TRG approach presents pairwise relation as a component of a learnable adjacent matrix to address the ambiguous temporal state relations of a video sequence. Additionally, the multi-head temporal relation graphs which discover temporal relations on a multi-scale range based on the adjacent matrix are constructed in our approach. Specifically, rather than convolving solely on the temporal sequence in the consecutive order, our proposed TRG approach defines specific relation graphs to explore the temporal relations of activity in both short-term and long-term temporal dependencies. Furthermore, the temporal instances in the graph have multi-relation between each other, like causality or adjacency, etc. Therefore, the multi-head temporal adjacent matrix is devised to investigate such relations. Subsequently, temporal relation reasoning can be performed to deduce multi-kinds of relation for an activity. To further exploit the temporal relation in the reasoning graph and investigate a semantic meaning of those relations, our proposed TRG approach designs a relation aggregator to fuse those multi-kinds of relation. The whole activity recognition procedure in our proposed TRG approach is illustrated in Figure \ref{fig:model}. Specially, ConvNets for spatial feature extraction of sampling video frames, our proposed TRG approach for temporal relation reasoning and the final fully connection for activity classification. Note that, in later experiment, we prove that our proposed TRG is flexible to be plugged into any stage of the off-the-shelf architectures, e.g., ResNet \cite{he2016deep} and Inception \cite{ioffe2015batch}.
    
    To further enhance the spatial-temporal features learning for activity recognition, we concatenate the spatial features extracted by ConvNets with the semantic temporal relation features. In this way, we can extract the spatial features of each frame and capture long-range temporal relation in videos. We conduct extensive experiments on three large-scale data sets: Something-Something V1 \cite{goyal2017something}, V2 \cite{mahdisoltani2018fine} and Charades \cite{sigurdsson2016hollywood}. Both three datasets are extremely challenging, even for a human, as we cannot infer the activity by only the object or background in the frame. The experiments on the two datasets have demonstrated the significant improvements over state-of-the-art approaches and the importance of temporal reasoning in activity recognition.
     
    Our major contributions are summarized as follows:
    \begin{itemize}
        \item We propose a novel temporal reasoning graph module for activity recognition which is the first attempt at temporal relation reasoning with graph, to our best knowledge.
        \item We construct multi-head graph representation for multi-kinds temporal relation reasoning of a video sequence with a variant span and scale between the sequence in a long-range video.
        \item  A semantic aggregator is developed to learn the importance of sequence state in different graphs and fuse the multi-kinds temporal relation features.
        % \item The results show the superiority of the proposed model in complex environments activity recognition.
    \end{itemize}
    
    The rest of this paper is organized as follows: we briefly introduce related work in activity recognition literature, visual relation reasoning as well as graph convolution networks in Section \ref{sec:related}. In section \ref{sec:method}, the proposed temporal reasoning graph (TRG) is elaborated in detail. Section \ref{sec:experiments} presents the experimental results and further analysis of the model and results. Finally, the conclusion is given in Section \ref{sec:conclusion}.

\section{Related Work}
\label{sec:related}

    \textbf{Activity Recognition.} Activity recognition has been widely studied in recent years. An active research which devotes to the design of deep networks for video representation learning has been trying to devise effective CNNs architectures \cite{karpathy2014large, varol2018long, tran2017convnet, varol2018long, donahue2015long}. Karpathy et al.\ \cite{karpathy2014large} attempted to design a deep network that stacks CNNs based frame-level features in a fixed size and then conduct spatiotemporal convolutions for video-level features learning. However, the results were not satisfying, which implied the difficulty of CNNs in capturing motion information of the video. Later, many works in this genre leverage CNNs trained on frames to extract low-level features and then perform high-level temporal integration of those features using pooling \cite{wang2018video, wang2018learning}, high-dimensional feature encoding \cite{Girdhar2017ActionVLAD, diba2017deep}, or recurrent neural networks \cite{donahue2015long, wu2015modeling, varol2018long, yue2015beyond}. To explore long-term temporal relationships of video for learning a more robust representation, recently, the convolution neural network and long short-term memory (CNNs-LSTM) frameworks \cite{donahue2015long, wu2015modeling} were applied by stacking LSTM network to connect frame-level representation. They do have yielded an improvement for modeling temporal dynamics of convolution features in videos. However, this genre using CNNs as an encoder and RNN as a decoder of the video would lose low-level temporal context which is essential for action recognition. These works implied the importance of temporal information for action recognition and the incapability of CNNs to capture such information. To exploit the temporal information, some studies resorted to the use of the 3D convolution kernel. Another efficient way to extract temporal features was to precomputing the optical flow using traditional optical flow estimation methods and training a separate CNNs to encode the precomputed optical flow, which is a kind of escape from temporal modeling but effective in motion features extraction. There are still several important issues with existing CNNs for action recognition: 1) CNNs has limited capacity for learning long temporal dependency, 2) it's difficult for CNNs to capture the temporal transformation with complex physical properties. To address these issues, we propose an efficient unit that applies graph convolution networks to learn temporal relation, which is much more efficient than convolving dense frames. Meanwhile, the model can construct a temporal graph for representing temporal relation. It flexibly incorporates temporal reasoning and spatial transformation with existing architectures.

    \textbf{Visual Relation Reasoning.} Reasoning about the relation between instance over time in the video is critical for activity recognition \cite{zhou2018temporal}. In addition, modeling relations between vision objects have become a popular problem in computer vision \cite{zhang2017visual, qi2018attentive, xiong2019visual, wang2019linkage}. 
    The most straightforward visual reasoning task is object relation reasoning in CLEVR benchmark \cite{johnson2017clevr}, and significant efforts have been devoted to a variety of traditional visual tasks with pairwise relationship reasoning. In the face clustering problem, several works attempted to modeling pair-wise relationships for face graph generation \cite{wang2019linkage, yang2019learning}. To reason between different instances in visual question answering tasks, Xiong et al.\ \cite{xiong2019visual} proposed a graph matching module for investigating such relation. In sketch-based action recognition, several works showed that modeling interactions of sketch joint can achieve excellent performance \cite{si2019an, shi2018non-local, stgcn2018aaai}. Here, we show that explicitly exploiting the various and multi-scale temporal relations in videos can boost activity recognition accuracy.

    \textbf{Graph Convolution Networks (GCNs).} GCNs is a powerful tool for modeling the graph instance relation. Spatial based GCNs \cite{zhou2018graph} which is the generalization of CNNs to graphs and perform manually-defined convolution on the graph is good at dealing with graph-structured data. Due to their convincing performance and high interpretability of modeling object relationships, GCNs has been widely applied in many computer vision task which needs to explore the relation of different vision instance. In terms of applications, existing works has led to considerable performance improvement by using GCNs in traditional computer vision tasks \cite{chen2018graph-based}, for example, skeleton-based action recognition \cite{stgcn2018aaai, shi2018non-local}, link prediction \cite{wang2019linkage, yang2019learning}, semi-supervised classification \cite{kipf2017semi-supervised}, hashing \cite{zhou2018hash, shen2017deep, shen2015supervised}, person-reid \cite{shen2018person}, and multi-label image recognition \cite{chen2019multi-label}, and etc.
    
    In our work, we exploit GCNs that is built by stacking multiple layers of graph convolutions with a multi-head adjacent matrix to capture temporal relations at multiple time scales. Moreover, our proposed method explicitly models the temporal interactions by building a temporal reasoning graph, which can be inflexible injected to the existing backbone. We show the temporal reasoning graph can efficient reason between temporal semantic instance for gaining activity classification accuracy.
    
\section{Proposed Method}
\label{sec:method}
    In this section, we illustrate the framework of our proposed architectures showed in Figure \ref{fig:model}, i.e., we will give detailed descriptions of how we build the temporal graph to investigate temporal relation. Firstly, the definition of the problem is given. Secondly, the detail of the construction of the temporal relation graph is described. Thirdly, the way how we implement convolution through the temporal graph is described in detail. Fourthly, the way we aggregate the semantic meaning of the multi-head temporal graph we constructed in Section \ref{subsec:graph} is presented. Finally, the way we model the spatiotemporal features and classify the activity is introduced.

\subsection{Problem Definition}
    Formally, we have extracted a sequence of features in a video as $\mathbf{X} = \left\{ x_1, x_2, \ldots, x_T \right\}$, where $T$ denotes the number of sequence or time of the video. If not specified, $x_i \in \mathbb{R}^{C \times H \times W}$ represents a frame feature map, $C$ is the feature channels, and $H, W$ represent the feature height and width, respectively. Here, we will apply the graph convolution neural network to explore the relation between temporal sequence $\mathbf{X}$ for activity recognition. Let $\mathcal{G} = \left ( \mathcal{X}, \mathcal{E} \right)$ denote a graph of temporal sequence, where $\mathcal{X}$ is the set of $T$ video frame object and $\mathcal{E}$ is the set of temporal relation edges between the video sequence. The neighbor set of a node $x_{i}$ is $\mathcal{N}\left ( x_{i} \right)$, and the nodes at adjacent time steps are connected with the temporal edge. The time series of a video can be easy obtained, but the semantic meaning of the edge $e = \left ( x_{i}, x_{j}\right ) \in \mathcal{E}$ between the node is still ambiguous. Attention mechanisms have been proven as a powerful tool in deep learning studies. We exploit the attention mechanism to investigate the semantic meaning of the edge $e$. More details about how we mine temporal relation will be elaborated in Section \ref{subsec:graph}.
    
\subsection{Temporal Relation Graph Construction}
\label{subsec:graph}
    In our work, we apply the temporal relation graph to form a hierarchical representation of the video sequence. We define the pairwise temporal relation and measure the similarity of the temporal feature to construct the temporal relation graph. We use the definition of temporal relation function as:
    \begin{equation}
        \centering
        e_{ij} =  g_{\theta}\left (x_{i}, x_{j} \right ),
        \label{equ:a}
    \end{equation}
    where $g_{\theta}$ is the similarity function of different frames. Note that $g_{\theta}$ has many formats, typically expressing as following:
    (1) \textbf{sum:} $g_{\theta} \left( x, y \right) = V^{T} tanh\left(x + y \right)$;
    (2) \textbf{dot product:} $g_{\theta} \left( x, y \right) = x^{T}y$; and (3) \textbf{bilinear:}
    $g_{\theta} \left( x, y \right) =  x^{T}\mathbf{W_1}y$. More specifically, two sequence objects that one of the object instances may be predicted by the other one will have close connection and a high confidence edge.

    In the temporal relation graph, we connect pairs of the semantically related frame features together. In order to obtain sufficient expressive power to transform the input frame feature into higher-level features, at least one transformation function which can project the feature sequences into a space for similarity measure is required. Thus, the new temporal relation function is:
    \begin{equation}
      e_{ij} = g_{\theta}\left (\mathbf{W}x_{i}, \mathbf{W}x_{j} \right ),  
    \end{equation}
    where $\mathbf{w}$ is the features transformation function parameter which can be learned via backpropagation. Therefore, a shared transformation, parameterized by a weight variable $\mathbf{W}$, is applied to every frame feature. By adding such transformation function, we could learn the adjacent matrix which represents the correlations between different temporal feature across the frame of the temporal graph at each single feedforward step.

    After computing the correlations coefficient of the adjacent matrix, we perform normalization across each row of the adjacent matrix for easily comparable across different temporal features. We adopt the softmax function for normalization as:
    \begin{equation}
        a_{ij} = softmax(e_{ij}) = \frac{\exp \left(e_{ij} \right)}{\sum_{k =1}^{T} \exp \left(e_{ik}\right)}.
        \label{equ:coeff}
    \end{equation}
    
\begin{figure}[h]
    \centering
    \includegraphics[width=0.9\columnwidth]{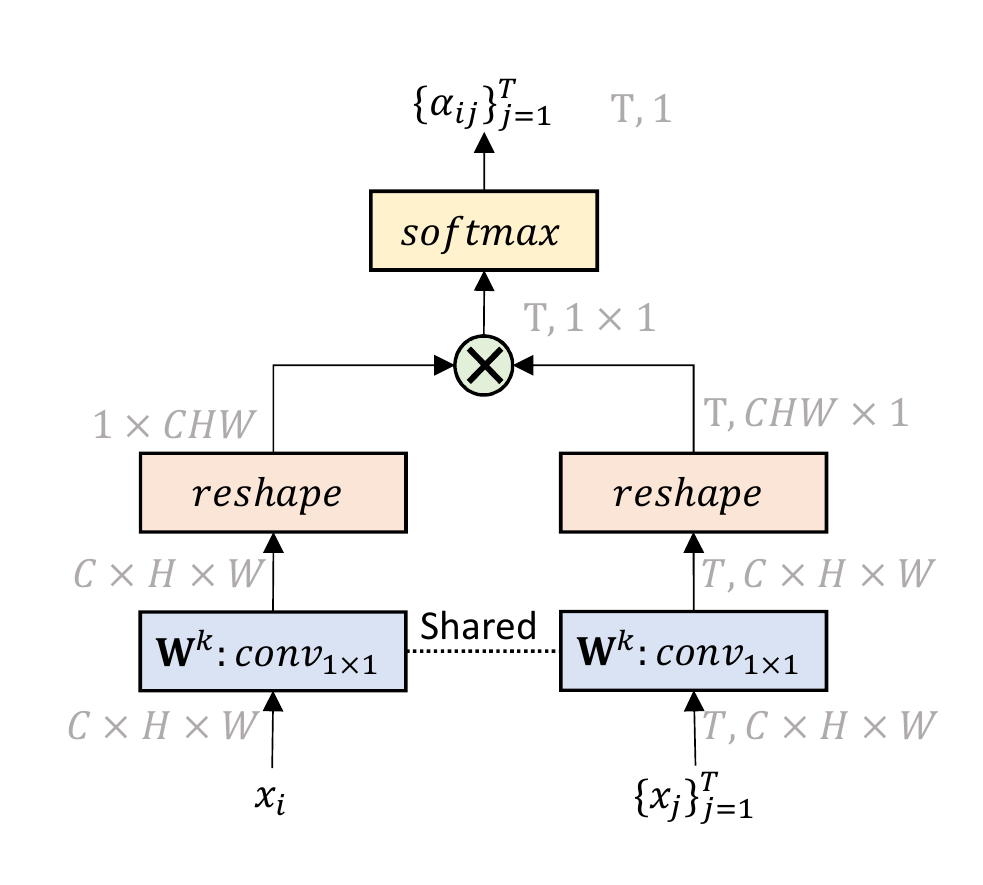}
    %\vspace{-0.1in}
    \caption{An illustration of the temporal similarity calculation operation in the adjacent matrix. The feature in a video sequence is shown in the tensor format. The transformation weight is implemented by $1\times1$ convolution kernel, and ``$\otimes$'' denotes matrix multiplication. All the features are convolved by the convolution kernel after proper reshaping. The temporal similarity is computed by the softmax operation implemented after the features multiplication.}
    \label{fig:module1}
    %\vspace{-1.1em}
    \end{figure}

    The learnable adjacent matrix $\mathbf{A}$ expresses as $\mathbf{A}=(a_{ij})_{T \times T}$. Here, to update the similarity measure of temporal relation at each step, we implement the adjacent matrix learning block by a single-layer feedforward neural network. An implementation example of element calculation in the adjacent matrix learning block is illustrated in Figure \ref{fig:module1}.
    
    %\vspace{-1.5em}

    We should notice that the temporal-based neighbors of each node may have many kinds of relations which show different importance in learning temporal feature. In order to obtain sufficient representational power to capture the underlying relation of the temporal sequence, we design a mechanism, named \textit{multi-head adjacent matrix}, to explore different semantic attributes.
    
    \textbf{Multi-head adjacent matrix.} Inspired by transformer \cite{vaswani2017attention} and graph attention networks \cite{velickovic2018graph}, we construct the multi-head adjacent matrix to explore the power of temporal relation and stabilize the learning process of sequence object correlation coefficients. Specifically, $N$ independent coefficient updating procedure execute the temporal relation function of Equation \ref{equ:coeff}:
    \begin{equation}
        a_{ij}^{k} = softmax\left(e_{ij}^{k}\right) = softmax\left( g_{\theta}\left( \mathbf{W}^{k}x_{i}, \mathbf{W}^{k}x_{j} \right) \right),
        \label{equ:node_weight}
    \end{equation}
    where $k \in \left\{ 1, 2, \ldots, N \right\}$. All the video temporal graph adjacent matrix denote as $\mathbf{A^{\prime}}=\left\{ \mathbf{A}^{1}, \mathbf{A}^{2}, \dots, \mathbf{A}^{N} \right\}$ , where $\mathbf{A^{k}}=(a_{ij}^{k})_{T \times T}$.
    
\subsection{Temporal Graph Convolution Neural Network}
    Once the temporal graphs are built, we can exploit graph convolution for temporal relation reasoning. Temporal graph convolution takes a video sequence as input, performs computations over the sequence, and returns a new sequence, i.e., the graph convolutions allow us to compute the response of a node based on its neighbors deﬁned by our multi-head graph. So, for a specific target object $x_{i}$ in the video sequence, it aggregates features from all neighbor objects according to the edge weight defined as previous. 

    To exploring temporal reasoning on the temporal relation graph, we apply the Graph Convolution Networks proposed in \cite{velickovic2018graph} to process the frame feature. The outputs of the GCNs are updated features of each frame node, which can be aggregated together for video classification. If we perform multi-head adjacent matrix updating mechanisms, the next state sequence feature will be:
    \begin{equation}
        z_{i}^{n} = \sigma\left( \sum_{j=1}^{T}a_{ij}^{n}\mathbf{W^{n}}x_j\right),
        \label{equ:node_feature}
    \end{equation}
    where $z_{i}^{n}$ is output semantic feature map reasoning through graph, $\sigma\left( \cdot \right)$ is typically nonlinear activation function $relu$ and $\mathbf{W^{n}}$ is transformation matrix implemented by standard convolution in feature domain. Note that the filter weight $\mathbf{W^{n}}$ in each graph is shared everywhere on feature sequences, because it is irrelevant to the location of the feature sequences. So after constructing multi-head temporal relation adjacent matrix $\mathbf{A}$ and apply graph convolution, we can obtain $N$ kinds of sequence state feature, denoted as $\left\{ z_{i}^{1}, z_{i}^{2}, \ldots, z_{i}^{N} \right\}$.
    % we employ aggregation function, like [] GraphSAGE,
    % \begin{equation}
    %     y_{i}= \sigma\left( aggregator\left(\sum_{t=1}^{K} \sum_{t=1}^{T}a_{ij}^{k}\mathbf{W}^{k}x_j\right) \right).
    % \end{equation}
    
\subsection{Multi-head Temporal Relation Aggregator}
    The multi-head adjacent matrix is exploited to discover the multi-kinds of relations between the sequence nodes. The main reason for designing the multi-head temporal relation aggregator is to reflect the different importance of the semantic relation. In short, the aggregator which exploits a mechanism similar to self-attention for temporal graph pooling. 

\begin{figure}[h]
    \centering
    \includegraphics[width=0.9\columnwidth]{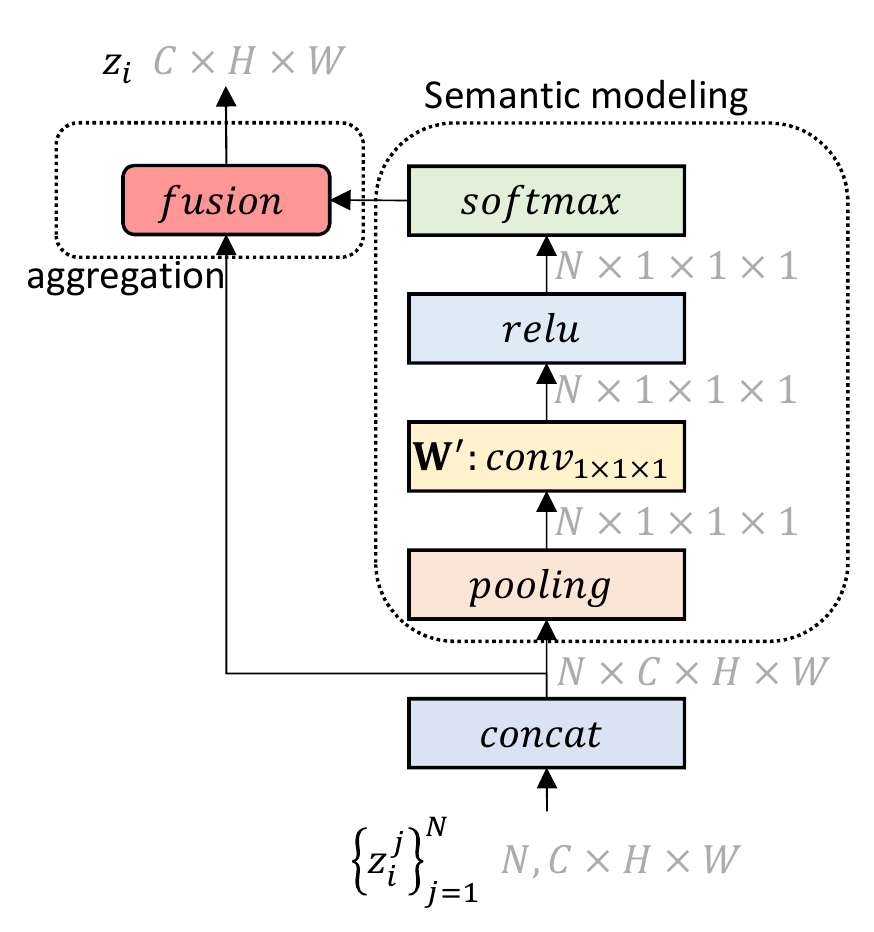}
    \vspace{-0.1in}
    \caption{The detail operation of the multi-head temporal relation aggregator. The semantic modeling mainly aims at extracting comprehensive information of the feature sequences in heterogeneous graphs. The aggregation is exploited to fuse the feature sequences with the global semantic modeling information.}
    \label{fig:module2}
    \vspace{-1em}
    \end{figure}

    Generally, every sequence instance in a heterogeneous graph contains multiple types of semantic information, i.e., semantic-specific sequence features extracted from one graph can only reflect temporal relation from one aspect. To learn a more comprehensive video feature, we need to fuse multiple semantics which can be revealed by multi-head adjacent matrix. To address the problem of semantic fusion in a heterogeneous graph, we propose a multi-head temporal relation aggregator to automatically learn the importance of sequence state in different graphs and fuse them. The operation of the temporal relation aggregator is illustrated in Figure \ref{fig:module2}. 
    % To significantly investigating the relation of multiply sequence frame, we extend the range of neighborhood . Similarity like convolution kernel in CNNs, we construct multiply frame features relation.
    To investigate the importance of different sequence states $z_{i}^{k}, k \in {1, 2, \ldots, N}$ which are updated based on different temporal graph adjacent matrix, we define an aggregator function as following to automatically learn the importance,
    \begin{equation}
        z_{i} = f_a \left( z_{i}^{1}, z_{i}^{2}, \ldots, z_{i}^{N} \right).
    \end{equation}

    To implement the function $f_a$, inspired by squeeze-and-excitation networks \cite{hu2018squeeze-and-excitation}, we first use global pooling as follows to extract the global semantic meaning of each state in the different graph,
    \begin{equation}
        \left( z_{i}^{\prime 1}, z_{i}^{\prime 2}, \ldots, z_{i}^{\prime N} \right)=pooling\left( z_{i}^{1}, z_{i}^{2}, \ldots, z_{i}^{N} \right),
    \end{equation}
    where $z_{i}^{\prime k} \in \mathcal{R}^{1\times1\times1}$. Then, the importance of each sequence state, denoted as $\beta_j$, is shown as follows:
    \begin{equation}
        \beta_j = relu\left( \mathbf{W^{\prime}}z_{i}^{\prime j} \right),
    \end{equation}
     the weight for each $z_{i}^{j}$ can be obtained by normalizing the above importance coefficient $\beta_{j}$ through all temporal graphs using the softmax function,
     \begin{equation}
         \beta_{i}^{\prime} =  \frac{\exp \left(\beta_{i} \right)}{\sum_{j =1}^{N} \exp \left(\beta_{j}\right)}.
         \label{equ:sem_weight}
     \end{equation}

     With the learned weight as coefficients, we can fuse these semantic-specific temporal features $z_{i}^{k}$ to obtain the final feature $z_{i}$ as follows:
     \begin{equation}
         z_{i} = \sum_{n=1}^{N}\beta_{n}^{\prime} \cdot z_{i}^{n}.
         \label{equ:agg_feature}
     \end{equation}
     
     Finally, the output of a video sequence features from the temporal graph convolution are aggregated. With the designed aggregator, the video sequence features after processed contain the semantic meaning of the multi-head temporal graph. For simplicity, the temporal graph convolution together with multi-head temporal relation aggregator operation are summarized as follows:
    \begin{equation}
        \mathbf{Z} = f_a \left( \sigma \left( \sum_{k=1}^{N} \mathbf{A}^{n}\mathbf{X}\mathbf{W}^{n} \right) \right),
        \label{equ:sem_spe}
    \end{equation}
    where $\mathbf{A}^{n}$ represents one of the multi-head adjacency graphs with $T \times T$ shape, and $\mathbf{X}$ is the input features map of all the sequence in the graph with shape $T \times C \times H \times W$. $\mathbf{W}^{n}$ is the spatial transformation matrix. Here, for computation simplicity, we exploit convolution kernel $conv_{3\times3}$ with one padding and one stride to represent $\mathbf{W}^{n}$, so $\mathbf{W}^{n}$ is a matrix of the layer with shape $C \times C \times 3 \times 3$. Note that the transformation matrix $\mathbf{W^{n}}$ in each graph are shared everywhere on feature sequences because it's irrelevant to the location of the feature sequences. The final output features map $\mathbf{Z}$ still is a tensor with shape $T \times C \times H \times W$. The temporal graph convolution operation and temporal relation aggregator can be stacked into multiple layers.
    
    \vspace{-1.1em}

\subsection{Spatial and Temporal Modeling}
    ConvNets has been proved a powerful tool for capturing the basic visual concept in recent years \cite{simonyan2014very, he2016deep, ren2017faster}, but it has a poor capacity of understanding the relation and importance of those spatial units in an image or a sequence for classification. From above, the temporal graph convolution is designed to extract such sequence relations of the spatial features. So we apply residual connection operation \cite{he2016deep} to fuse the spatial features extracted by ConvNets with the temporal relation features extracted by temporal graph convolution for activity recognition.
    
    After the video sequence spatial features $\mathbf{X}$ extracted by ConvNets are obtained and the temporal relation features $\mathbf{Z}$ are extracted by temporal graph convolution, we begin to model the spatial and temporal feature extraction by connecting the $\mathbf{X}$ and $\mathbf{Z}$ as follows:
    \begin{equation}
        \mathbf{H}=\sigma \left( \mathbf{X} + \mathbf{Z}\right),
        \label{equ:spa_tem}
    \end{equation}
    where ``$+$'' is the residual concatenation operation, $\mathbf{H}$ is the concatenated spatial-temporal features, followed by a non-linearity function $\sigma\left( \cdot\right)$, typically \texttt{ReLU}. The process of spatial and temporal modeling procedure is summarized as Algorithm \ref{alg:st_model}.
    
    %%%%%%%%%%%%%%%%%%%%%%%%%%%% algorithm %%%%%%%%%%%%%%%%%%%%
    \begin{algorithm}[ht]
	%     \SetAlgoNoLine % 
	\SetKwInOut{Input}{\textbf{Input}}\SetKwInOut{Output}{\textbf{Output}} % 
	
	\Input{
		The spatial features sequence of a video $\mathbf{X} = \left\lbrace x_1, x_2, \ldots, x_T\right\rbrace$.
	}
	\Output{
		The final spatial and temporal representation $\mathbf{H}$. 
	}
	\BlankLine
	\For {$ x_i  \in \left\lbrace x_1, x_2,\ldots,x_T \right\rbrace $}{
		\For {$k=1 \ldots N$}{
			Calculate the graph node relation weight $a_{ij}^k$ using Eq. \ref{equ:node_weight}\;
			Calculate the graph node feature $z_i^k$ using Eq. \ref{equ:node_feature}\;
		}
		%sss \;
	}
	%Concatenate the graph node relation weight to form the adjacent matrix set $\mathbf{A\prime} = \left\lbrace \mathbf{A}_1, \mathbf{A}_2, \ldots, \mathbf{A}_T \right\rbrace$ \;
	
	\For{ $i=1 \ldots T$}{
	    \For{$z_i^j \in  \left\lbrace z_i^1, z_i^2, \ldots ,z_i^N \right\rbrace$}{
	        Calculate the semantic weight $\beta_j$ of $j^{th}$ node in $i^{th}$ feature sequences using Eq. \ref{equ:sem_weight}\;
	    }
	    Calculate the aggregate feature $z_i$ using Eq. \ref{equ:agg_feature}\;
	}
	Concatenate the semantic-specific temporal feature $\mathbf{Z} = \left\lbrace z_1, z_2, \ldots, z_T \right\rbrace$\; 
	%Calculate task-specific loss \;
	Fuse spatial and temporal feature to form $\mathbf{H}$ using Eq. \ref{equ:spa_tem}\;
	Back propagation and update parameters in TRG\;
	\Return{$\mathbf{H}$. }
	\caption{The process of spatial and temporal modeling procedure in our proposed TRG.}
	\label{alg:st_model}
    \end{algorithm}
    
\subsection{Activity Classification with Temporal Reasoning Graph}
    The temporal reasoning graph provides a solid unit to work with existing architectures. As illustrated in Figure \ref{fig:model}, the input is processing via spatial transformation and temporal graph reasoning to form the video representation, and then the representation is fed into a classifier to generate activity label. To this end, the whole framework can be trained in an end-to-end manner. For single-label activity recognition tasks, we apply cross-entropy loss when training, while binary sigmoid loss is used for multi-label activity recognition task. The loss function is given as:
    % \begin{equation}
    %     L = \begin{cases}
    %      \left(-s_{i, g} + log \sum_{j=1}^{C}exp(s_{i, j}) \right), \\ 
    %      \quad \quad \quad \quad \quad \quad \quad \quad \quad \quad \quad  \text{if single-class classification;} \\
    %     \sum_{j=1}^{C} -\left( y_{i, j} log(\delta(s_{i, j})) + (1-y_{i, j})log(1-\delta(s_{i, j})) \right), \\ 
    %      \quad \quad \quad \quad \quad \quad \quad \quad \quad \quad \quad \text{if multi-class classification;}
    %     \end{cases},
    % \end{equation}
    \begin{equation}
        L = \begin{cases}
        \mathcal{L}_1, & \text{for single-class classification;} \\
        \mathcal{L}_2, & \text{for multi-class classification.}
        \end{cases}
    \end{equation}
    where $\mathcal{L}_1 = \left(-s_g + log \sum_{j=1}^{C}exp(s_j) \right)$, $s_j$ is the confidence scores of class $j$, $g$ is the ground true label and $C$ is the number of classes, and $\mathcal{L}_2 = \sum_{j=1}^{C} -\left( y_j log(\delta(s_j)) + (1-y_j)log(1-\delta(s_j)) \right)$, $\delta()$ is \texttt{sigmoid} function and $y_i=1$ when classes $i$ is the ground true label of the sample and 0 otherwise.
    
    Using the above objective function, the parameters of the whole framework are optimized with the standard mini-batch stochastic gradient descent (SGD) algorithm with Nesterov momentum which is widely adopted in activity recognition task \cite{wang2016temporal, zhou2018temporal, wang2018videos}.

\section{Experiments}
\label{sec:experiments}

    To evaluate the effectiveness of our temporal reasoning graph, we perform extensive experiments on three benchmark datasets for activity recognition: Something-Something V1 \cite{goyal2017something}, V2 \cite{mahdisoltani2018fine} and Charades \cite{sigurdsson2016hollywood}. We first introduce the two datasets and implementation detail. Then, we compare our method with existing methods. Afterward, we conduct ablation studies to dig into the effect of the component and factor. In addition, we further analyze the temporal scaling strategy, where to place our temporal reasoning graph and recognition performance with different backbone. Finally, we provide the visualization of class activation map (CAM) \cite{zhou2016learning} of the intermediate features and apply t-SNE \cite{dermaaten2008visualizing} to visualize the distribution of feature representation learned by our model.

\subsection{Datasets Description}
    \textbf{Something-Something-V1} \cite{goyal2017something}: The dataset is a large collection of densely-labeled video clips that allow machine learning models to develop a fine-grained understanding of basic actions. It contains 198,499 short video clips across 174 labels of simple textual descriptions based on templates, 86,017 of which are training videos, 11,522 are validation videos and 10,960 are testing videos. Each video has a duration ranging from 2 seconds to 6 seconds. The inference of the video in this dataset requires extracted features that are capable of representing the physical relation of the objects. 
    
    \textbf{Something-Something-V2} \cite{mahdisoltani2018fine}: The dataset is twice as many videos as V1, collected by workers with humans performing pre-defined basic actions of everyday objects. Each video in the dataset equips with object annotations in addition to the video label and has a duration ranging from 2 to 6 seconds. In total, It has 318,572 annotations involving 30,408 unique objects and contains 220,847 videos across 174 class, with 168913 of which are training videos, 24,777 are validation videos and 27,157 are testing videos, i.e., it is split into train, validation, and test-sets in a ratio of 8:1:1.
    
     \textbf{Charades} \cite{sigurdsson2016hollywood}: The dataset is composed of 9,848 videos of daily indoors activities with an average length of 30 seconds, involving interactions with 46 objects classes in 15 types of indoor scenes and containing a vocabulary of 30 verbs leading to 157 action classes. Each video in this dataset is annotated by multiple free-text descriptions, action labels, action intervals and classes of interacting objects. 267 different users were presented with a sentence, which includes objects and actions from a fixed vocabulary, and they recorded a video acting out the sentence. In total, the dataset contains 66,500 temporal annotations for 157 action classes, 41,104 labels for 46 object classes, and 27,847 textual descriptions of the videos. Following the standard split, it has 7,986 training video and 1,863 validation video.
    
    These datasets provide us a large number of samples to investigate our model in video activity understanding and commonsense reasoning for daily human activities.
    
\subsection{Implementation Detail}
    \textbf{Training.} We adopt the training strategy proposed in TSN \cite{wang2016temporal}, like data enhancement and partial batch normalization, etc. The frames sampled from a video were first input to 2D or 3D ConvNets for spatial feature extraction. Then we apply the temporal reason graph to learn the high-level semantic meaning based on the sequence spatial features. All the 2D or 3D spatial feature extraction processes are originated from \cite{wang2016temporal, wang2018non, wang2018videos} for a fair comparison.
    
    We employ the PyTorch framework \cite{paszke2017automatic} in this paper for Networks building, and all networks are trained on two GeForce GTX Titan X GPU with a total 24G memory. All the input images are resized to 224 $\times$ 224 with the backbone of Inception \cite{ioffe2015batch} and ResNet \cite{he2016deep} and 299 $\times$ 299 with the backbone of Inception-V3 \cite{szegedy2016rethinking} followed by the dataset processing strategy of \cite{wang2016temporal}. The network is trained with a mini-batch stochastic gradient descent optimizer for model training, and the initial learning rate here is 0.001 which will reduce by a factor 10 after 50 epochs. It has a decay rate of $5 \times 10^{-4}$, and momentum 0.9 to update Network parameters. The whole training procedure takes 100 epochs. For Something-Something V2, the epoch number is halved because the duration of its videos is shorter. All the learnable Parameters in the temporal graph are implemented by a convolution layer followed by a batchnorm \cite{ioffe2015batch} and relu layer \cite{glorot2011deep}.
    
    % We take different number of video frames of Charades and Something-Something datasets when training and investigate its' effect on the model performance at \ref{S:4.3}.
    
    \textbf{Test.} When comparing with previous state-of-the-art models, we followed the rule that uses the same number of frames for both training and testing to make a direct comparison. Following the common practice, the inference process is conducted by sampling 10 clips and 2 clips from a video along its temporal axis for Charades and Something-Something dataset respectively and the prediction scores are averaged over whole clips. By the way, we also rescale the shorter side to 256 pixels for each frame while maintaining the aspect ratios.

\subsection{Evaluation Protocols}
    Every video in Something-Something-V1 and V2 datasets is assigned to a single classes and the distribution of classes over the test set is almost uniform. The ground truth labels of the videos are $g_i, i=1, \ldots, N$, where $N$ is the number of samples in the test set. Following \cite{goyal2017something, wang2018videos, zhou2018temporal}, we apply $top1$ and $top5$ precision to evaluate the performance of several methods in this datasets. The methods produce a list of at most 5 action classes labels, $l_{i, j}, j=1, \ldots, 5$, in the descending order of confidence for each video $i$. The $topk$ ($k=1, \ldots, 5$) precision is defined as: $topK = \frac{1}{N} \sum_{i}^{N}\sum_{j=1}^{K} I(l_{i, j}, g_i)$, where $I(x, y)=1$ if $x=y$ and 0 otherwise. Under this circumstance, the $top1$ precision is when $K$ set to 1 and $K$ set to 5 for $top5$.
    
    As Charades is a multi-label action classification dataset, following \cite{sigurdsson2016hollywood, wang2018videos, zhou2018temporal}, we exploit the standard mean average precision (mAP) to measure the performance of several different methods. Here, mAP is defined as the mean of all classes' average precision (AP), where AP is formulated as $AP = \frac{1}{N}\sum_{k=1}^{K}P(k) \times rel(k)$, where $N$ is the number of positive samples in the test set, $P(k)$ is the precision of the top $K$ test samples, and $rel(k)$ is an indicator function equaling 1 if the item at rank $k$ is a positive sample, 0 otherwise.
 
% Comparison with state-of-the art 
\subsection{Compared Methods}
    We compare the performance of our TRG with the following state-of-the-art methods:
    \begin{itemize}
        \item \textbf{C3D} (ICCV 2017) \cite{goyal2017something}: A baseline model provided in Something-Something dataset exploits multi-layer 3D convolution \cite{tran2015learning} to evaluate the dataset.
        \item \textbf{MultiScale TRN} (ECCV 2018) \cite{zhou2018temporal}: An improved temporal aggregation method that applies multiple full connection layers to reasoning the relation of temporal features extracted by 2D-CNNs.
        \item \textbf{I3D} (CVPR 2017) \cite{carreira2017quo}: The methods inflate the 2D convolution filters to the 3D convolution filters, which can be transferred to many existing architectures.
        \item \textbf{NL} (CVPR 2018) \cite{wang2018non}: Non-local block can be plugged into an existing backbone to capture long-term dependencies. 
        \item \textbf{GCNs} (ECCV 2018) \cite{wang2018videos}: The model first applies faster rcnn to extract object features in the video and then use GCNs to reasoning the relation between those semantic objects.
        \item \textbf{ECO} (ECCV 2018) \cite{zolfaghari2018eco:}: An end-to-end architecture injects 3D CNNs at the top of 2D CNNs architecture.
        \item \textbf{TrajectoryNet} (NIPS 2018) \cite{zhao2018trajectory}:  Trajectory convolution, a replacement operation of temporal convolution, is proposed to integrate feature along the temporal dimension.
        \item \textbf{2-stream} (NIPS 2014) \cite{simonyan2014two}: The method designs a spatial and temporal stream for action recognition. The spatial stream extracts spatial feature from a single frame in a video, whilst the temporal stream extracts temporal feature from the pre-computed dense optical flow.
        \item \textbf{Asyn-TF} (CVPR 2017) \cite{sigurdsson2017asynchronous}: The model attempts to reason over various aspects of activity that includes objects, actions, and intentions.
    \end{itemize}

\subsection{Results and Discussion}

    \begin{table*}[!htb]
   %\small
   %\setlength{\tabcolsep}{4pt}
        %\centering
        \caption{Comparison of state-of-art methods on the Something-Something V1 and V2 datasets. We only report the top1 recognition accuracy using RGB frames as input without optical flow.}
        \begin{center}
        \begin{tabular}{lcccccccc}
        \toprule
         \multirow{2}{*}{Methods} & \multirow{2}{*}{Backbone}&  \multirow{2}{*}{\# Frames}& \multirow{2}{*}{\# Params}& \multirow{2}{*}{FLOPs}& \multicolumn{2}{c}{V1} & \multicolumn{2}{c}{V2}\\
         & & & & & top-1 & top-5 & top-1 & top-5\\
         \midrule
         C3D \cite{goyal2017something} ~ & VGG16 & 60 & 23.3M & 349.4G & 27.2 & - & 47.7 & 77.3\\
         MultiScale TRN \cite{zhou2018temporal} ~& Inception & 8 & 18.3M & 16.4G & 34.4 &- & 48.8 &- \\
         I3D~\cite{wang2018videos} & ResNet-50 & 32 & 28.0M & 153G & 41.6 & 72.2 & - & -  \\
         I3D + GCNs \cite{wang2018videos}  ~& ResNet-50 & 32 & 55.1M & 158G & 43.3 &  75.1 &- &-  \\
         NL I3D \cite{wang2018videos}  ~& ResNet-50 & 32 & 35.1M & 168G & 44.3 &  75.1 &- &-  \\
         NL I3D + GCNs \cite{wang2018videos}  ~& ResNet-50 & 32 & 62.2M & 303G & 46.1 &  76.8 &- &-  \\
         ECO$_{en}$ \cite{zolfaghari2018eco:} ~& Inception+3DResNet-18 & 92 & 150.0M & 267G & 46.4 & - &- &- \\
         TrajectoryNet \cite{zhao2018trajectory} ~ & S3DResNet-18 & 16 & - & - & 47.8 & - &- &-\\ \hline
         TRG~ & Inception & 8 & 14.1M & 16.2G & 38.5 & 68.4 & 51.3 &78.8 \\
         TRG~ & Inception & 16 & 14.1M & 32.1G & 45.9 & 74.9 & 56.7 &79.9 \\
         TRG~ & Inception & 32 & 14.1M & 64.5G &  47.5 & 80.2 & 58.3 & 86.2 \\
         TRG~ & Inception-V3 & 8 & 28.3M & 47.4G & 41.3 & 73.4 & 52.5 & 80.6 \\
         TRG~ & Inception-V3 & 16 & 28.3M & 94.6G & 47.2 & 78.9 & 59.2 & 86.4 \\
         TRG~ & Inception-V3 & 32 & 28.3M & 187.3G & \textbf{49.7} & 85.3 & 61.3  &\textbf{91.4}\\
         TRG~ & ResNet-50 & 8 & 28.4M & 33.6G & 41.5 & 74.7 & 53.8  &81.2\\
         TRG~ & ResNet-50 & 16 & 28.4M & 66.4G & 48.1 & 80.4 & 59.8  & 87.4\\
         TRG~ & ResNet-50 & 32 & 28.4M & 132.2G & 49.5 & \textbf{86.1} & \textbf{62.2}  & 90.3\\
         \bottomrule
        \end{tabular}
        \end{center}
        \vspace{-3mm}
        \label{tab:compare_smth}
    \end{table*}
    
    \begin{table}[!htb]
%   \label{tab:compare_cha}
   %\small
   %\setlength{\tabcolsep}{4pt}
    \caption{Comparison of state-of-art methods in terms of on the Charades dataset. ''$x/y$`` in the third column of input frames represents the input has $x$ samples RGB image and $y$ samples optical flow image. Other methods only take RGB modality into account.}
        \begin{center}
        \begin{tabular}{lcccccc}
        \toprule
         Methods & Backbone& \# Frames& mAP \\
         \midrule
         2-Stream~\cite{sigurdsson2017asynchronous} & VGG16 & 1/20 & 18.6 \\
         2-Stream +LSTM~\cite{sigurdsson2017asynchronous} & VGG16 & 1/20 & 17.8 \\
         Asyn-TF~\cite{sigurdsson2017asynchronous} & VGG16  & 1 & 18.3  \\
         MultiScale TRN~\cite{zhou2018temporal} & Inception & 8 & 25.2  \\
         I3D~\cite{carreira2017quo} & Inception & 32 & 32.9  \\
         I3D~\cite{wang2018non} & ResNet-101 & 32 & 35.5  \\
         NL I3D~\cite{wang2018non}  & ResNet-101  & 32 & 37.5  \\
         NL I3D + GCNs~\cite{wang2018videos} & ResNet-50 & 32 & 37.5  \\
         I3D + GCNs~\cite{wang2018videos} & ResNet-101 & 32 & 39.1  \\
         NL I3D + GCNs \cite{wang2018videos}  ~& ResNet-101 & 32 & 39.7   \\ \hline
         TRG~ & Inception & 8 & 28.4   \\
         TRG~ & Inception & 16 & 30.7  \\
         TRG~ & Inception & 32 &  33.5  \\
         TRG~ & Inception-V3 & 8 & 32.1  \\
         TRG~ & Inception-V3 & 16 &  35.1 \\
         TRG~ & Inception-V3 & 32 & 37.9 \\
         TRG~ & ResNet-50 & 8 & 32.9 \\
         TRG~ & ResNet-50 & 16 & 35.8 \\
         TRG~ & ResNet-50 & 32 & 38.4 \\
         TRG~ & ResNet-101 & 8 & 35.7 \\
         TRG~ & ResNet-101 & 16 & 37.9 \\
         TRG~ & ResNet-101 & 32 & \textbf{40.2} \\
         \bottomrule
        \end{tabular}
        \end{center}
        \vspace{-3mm}
        \label{tab:compare_cha}
    \end{table}

    The results on Something-Something \cite{goyal2017something} and Charades \cite{actorobserver, sigurdsson2016hollywood} are shown in Table \ref{tab:compare_smth} and \ref{tab:compare_cha} respectively. ``\# Frames'' in the table denotes the number of frames used as input by the method, whilst ``\# Params'' denotes the model parameters of the method, and ``FLOPs'' (floating point operations) is a type of measurement of the model computation complexity. For a fair comparison, we conduct a series of experiments with different input frames and different backbone to evaluate our model. For all of these datasets, we use the standard evaluation protocol provided by the authors. Note that we only exploit video frames as input without complementary information, like hand-crafted features-IDT \cite{wang2013action} or optical flow field.
    
    Table \ref{tab:compare_smth} shows all the results on Something-Something V1 and V2 validation set. Considering that they are fine-grained activity understanding datasets, in which deformation or motion serves as a crucial cue, those recent methods design their model from different aspects. But they didn't take the long-term temporal relation into account. Compared with Multiscale TRN which also attempt to reason about the temporal state of a video sequence, our work surpasses by a large margin 4.1\% on Something-Something V1 and 2.5\% on Something-Something V2 with the same experimental settings. The comparison with the work \cite{wang2018videos} which also apply graph convolution to reason the object proposal extracted by faster-rcnn \cite{ren2015faster} is impressive. Here, we gain 3.7\% accuracy improvement with this work. Since this approach considers the object-level relation in different frames, it first use faster-rcnn to extract object-level features, which may be harmful to their model setting if the model extracts negative object or missing some key object. Additionally, their computation cost is higher that our method. The margin may benefit from our multi-head adjacent matrix and semantic relation aggregator. Based on Inception-V3 architecture, our temporal reasoning graph widens the advantage over previous models considerably, bringing overall performance to 49.8\% on Something-Something V1 and 61.3\% on Something-Something V2. Compare with the most accurate method TrajectorNet which convolves through trajectory path, our method can still yield improvement. Although TrajectoryNet can work with a relatively small backbone to achieve a high performance, it has to apply MotionNet \cite{zhu2018hidden} to pre-produce trajectory as the performance of TrajectoryNet highly depends on the quality of the trajectory. Those results indicate that temporal reasoning with our graph model can capture semantic features for activity recognition. 
    
    In addition, the evaluation results on the Charades dataset, a multi-label activity recognition case, are shown in Table \ref{tab:compare_cha}. We present mAP values of those methods in this dataset. Our temporal reasoning graph achieves competitive recognition performance with the non-local inflated 3D CNNs (I3D) plus graph convolution model. Not only can our model get higher performance than \cite{wang2018videos}, but also the complexity of our model is very low. Since we need not detect objects from the frames, our model has a faster inference speed with only 0.3 seconds at a time for one video (32 frames).

    Table \ref{tab:compare_smth} and \ref{tab:compare_cha} also compares the inﬂuence of architecture and the number of sampled frames. As observing from the table, the accuracy degraded with fewer frames partly due to important parts of the action may be missing with fewer samples. However, training with more frames has a higher computation cost. The evaluation performance of TRG, even with just 8 frames, is still much better than most approaches in this literature, since our model takes into account the relationship between these 8 instants in the video, even if they are far apart.
    
\subsection{Further Analysis}
    In this subsection, we mainly explore the effectiveness of temporal reasoning graph block. To analyze the TRG in-depth, we evaluate the different factors which matter in action representation or be essential in our TRG.
    
\label{S:4.3}
    \noindent\textbf{Component Analysis of The Model.} To study the contribution of different model parts, we also train ablated versions of our model separately, i.e., without temporal reasoning in the video context. We first evaluate the effect of using the multi-head adjacent matrix for graph convolution in the Inception, Inception-V3 and ResNet-50 with I3D technology architectures as demonstrated in Table \ref{tab:compare_smth}. In the table, ``w/o temporal graph'' means without temporal graph, whilst ``temporal graph with'' means replacing temporal relation aggregator with a specific operation. Firstly, without temporal graphs, we apply temporal average pooling among features map as three baselines. In addition, for the 2D CNNs case, we also test plugging 3D convolution layer behind the end of each branch (i.e., \texttt{inception3a} to \texttt{inception5b} and \texttt{Mixed5b} to \texttt{Mixed7c}). Finally, for exploring the effect of the multi-head temporal relation aggregator. We replace the aggregator with an element-wise average operation and feature concatenation operation among the sequence depth dimension. All the experiments here are conducted with only 8 input frames.
    
    We can obtain that the reported baselines typically underperform the proposed model. The performance which yields improvement with our model demonstrates that temporal reasoning with multi-kinds and multi-scale sequence instance relation helps to capture fine grain action for recognition and boosts the performance. We speculate this is because the temporal graph considerably explores the semantic information of a video sequence and the temporal relation aggregator is able to select more informative relation features for activity recognition.

\begin{table}
   %\small
   %\setlength{\tabcolsep}{4pt}
        \caption{Ablation study of basic module of our model on Something-Something V2 dataset.}
        \begin{center}
        \begin{tabular}{lcccccc}
        \toprule
         Backbone & Module & top1 & top5\\
         \midrule
         \multirow{5}{*}{Inception} &  AvgPool w/o temporal graph & 37.3 & 68.9\\
         &3DConv w/o temporal graph & 42.4 & 73.2\\
         & temporal graph with concatenation & 48.5 & 77.4\\
         & temporal graph with element-wise Avg & 49.3 & 78.2 \\
         & TRG (full model) & \textbf{51.3} & \textbf{78.8} \\ \hline
         \multirow{5}{*}{Inception-V3} &  AvgPool w/o temporal graph & 39.5 & 70.1\\
         &3DConv w/o temporal graph & 45.1 & 76.5 \\
         & temporal graph with concatenation & 50.4 & 78.4\\
         & temporal graph with element-wise Avg & 51.2 & 79.2\\
         & TRG (full model) & \textbf{52.5} & \textbf{80.6}\\ \hline
         \multirow{4}{*}{ResNet-50} &  AvgPool w/o temporal graph & 46.7 & 74.1\\
         % &3DConv w/o temporal graph & - & - \\
         & temporal graph with concatenation & 50.6 & 77.6\\
         & temporal graph with element-wise Avg & 51.9 & 79.5\\
         & TRG (full model) & \textbf{53.8} & \textbf{81.2}\\
         \bottomrule
        \end{tabular}
        \end{center}
        \vspace{-3mm}
        \label{tab:abla-module}
\end{table}
    
\begin{figure}[h]
    \centering
    \includegraphics[width=0.9\columnwidth]{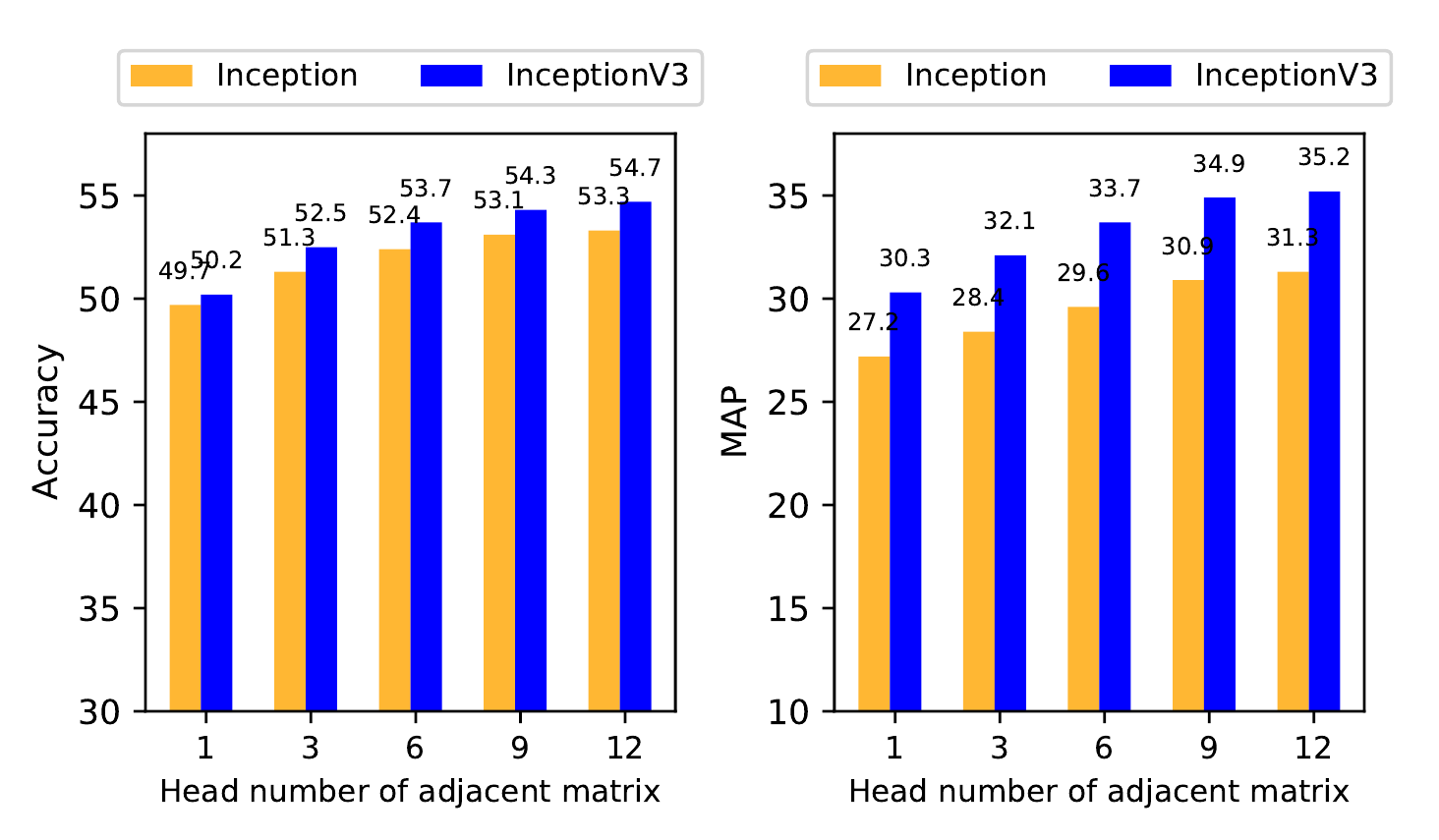}
    \caption{The performance of different head numbers of the adjacent matrix with Inception or Inception-V3 architecture on Something-Something V2 (left) and Charades (right) datasets.}
    \label{fig:head}
    \end{figure}

    \noindent\textbf{Head Number of The Graph.} To analyze the effect of the different numbers of the multi-head adjacent matrix on recognition performance, We perform a serial of experiments on Something Something V2 and Charades datasets with the different head numbers. Here, we only sample 8 frames for experiments. As shown in Figure \ref{fig:head}, we observe that building multiple multi-head temporal adjacent matrix lead to conspicuous gain compared with only one head adjacent matrix. The model is able to further boost accuracy from 49.7\% to 53.3\% with Inception backbone, 50.2\% to 54.7\% with Inception-V3 backbone on Something-Something V2, and improve the mAP from 27.2\% to 32.3\% with Inception backbone, 30.3\% to 35.2\% with Inception-V3 backbone on Charades dataset. The intuition behind this is that two temporal objects have different kinds of relations. When we deduce an action from a specific video, some kinds of relations may be enhanced and the others may be restricted by the relation aggregator. Although the adjacent relation weight values in the matrix learned with different head encode different semantic information, another observation that the recognition performance is saturated when keep increasing the number of the head indicated that the accumulation of the encoded semantic information with endless has little contribution and may cause confusion for relational reasoning.
    
\noindent\textbf{Place to Inject Temporal Reasoning Graph.} To figure out in which way the temporal reasoning graph helps in video context interaction, we conduct experiments of plugging temporal reasoning graphs in different places of the CNNs backbone. Furthermore, in this part, we will demonstrate how to push the temporal reasoning graph to the 2D or 3D CNNs backbone.
    
    Here, we will apply vanilla Inception as backbone-V3 for 2D CNNs case, whereas, ResNet-50 with I3D technique as backbone for 3D CNNs case. Theoretically, the temporal reasoning graph described above can be plugged at any level of the networks to enhance the temporal sequence interaction. In detail, we will study the effect of the different place where choose to build temporal reasoning graph. We push three graphs into the top or bottom of the multi-branch convolution networks. More specifically, we consider injecting the temporal reasoning graph into the output layers from \texttt{mixed\_5b} to \texttt{mixed\_5d} at bottom-level and from \texttt{mixed\_6a} to \texttt{mixed\_6c} for at bottom-level for Inception-V3. As a comparison, we consider processing output from \texttt{mixed\_7a} to \texttt{mixed\_7c} with our temporal reasoning graph for Inception network at the top-level. Similar processing will be applied to ResNet-50 from \texttt{res2\_2} to \texttt{res2\_4}, \texttt{res3\_2} to \texttt{res3\_4} or \texttt{res5\_2} to \texttt{res5\_4}.

\begin{figure}[ht]
    \centering
    \includegraphics[width=0.8\columnwidth]{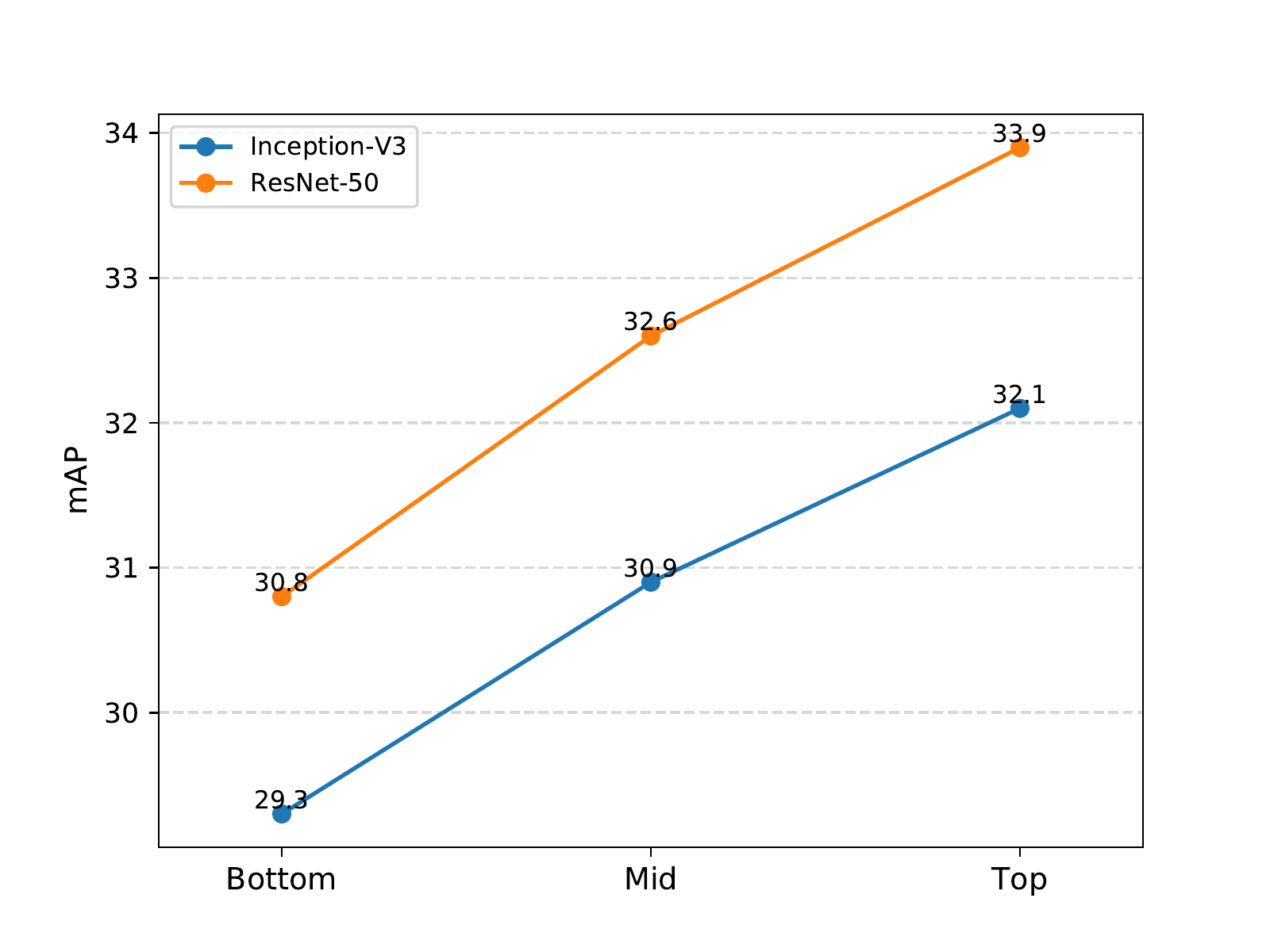}
    \caption{The performance on Charades dataset with different position to add temporal reasoning graph.}
    \label{fig:stage}
    \end{figure}

    Results are shown on Figure \ref{fig:stage} with 8 frames input case and the performance with top-level processing are higher. We speculate that output feature maps of late ConvNets layers contain abundant semantic information of the spatial objects. With the temporal reasoning of that semantic information, the global contextual good for activity representation is perfectly modeling for recognition.

    \noindent\textbf{Study on Temporal Scale.} The videos have a variable number of frames, and the activity in the video lasts seconds to even minutes in different datasets. To copy with the problem of action contextual information ranging from seconds to even minutes, we analyze the impact of different temporal stride and temporal spanning for sampling from the original video.  By the way, all the experiments were conducted with Inception, and ResNet-50, a 3D model processed the same with \cite{wang2018non}, architectures.
    
    The Uniformly sparse and global sampling strategy are exploited to select comprehensive frames from the entire video. Temporal pooling, except 3D CNNs case, process frames independently and their scores are aggregated only in the end. Consequently, the performance stays almost the same when they change the number of samples, which demonstrates that only with sparse sampling strategy does not really help to learn the long-range temporal context. Whereas, our model pays more attention to the temporal interaction of the video sequence. The performance indicates that our model really benefits from long-term temporal reasoning.
    
    By contrast, we also apply dense sampling strategy with a fixed temporal stride 4, which means that this strategy covers a temporal range of ~5 seconds at most in our experiments on Charades datasets (24 FPS). Uniformly spare sampling with large temporal striding, in fact, hurts the performance of 3D CNNs model. Temporal convolution kernel might not be suitable for long-range patterns, because long-range contextual patterns are more diverse and changeable, and include challenging scene cuts. On the other hand, large temporal striding with plugging temporal reasoning graphs steadily improves performance.
    
    Temporal reasoning sufficiently exploits the comprehensive information from the entire video or several seconds clip, since our approach modeling the relation of changeable pattern in a video sequence both with short and long-range.
    
    \begin{table}
   %\small
   %\setlength{\tabcolsep}{4pt}
        \caption{The performance of different video sampling strategy on Charades datasets.}
        \begin{center}
        \begin{tabular}{lcccccc}
        \toprule
         Sampling strategy & Methods & Backbone & \# Frames & mAP\\
         \midrule
         \multirow{8}{*}{Sparsely} &  Temporal pooling & Inception & 16 & 30.4\\
         & Temporal pooling & Inception & 32 & 31.6\\
         & TRG & Inception & 16 & 30.7 \\
         & TRG & Inception & 32 & 33.5 \\
         & Temporal pooling & ResNet-50 & 16 & 30.5\\
         & Temporal pooling & ResNet-50 & 32 & 32.3\\
         & TRG & ResNet-50 & 16 & 34.1 \\
         & TRG & ResNet-50 & 32 & \textbf{36.8} \\\hline
         \multirow{8}{*}{Densely} &  Temporal pooling & Inception & 16 & 30.1\\
         & Temporal pooling & Inception & 32 & 31.3\\
         & TRG & Inception & 16 & 29.8 \\
         & TRG & Inception & 32 & 32.7 \\
         & Temporal pooling & ResNet-50 & 16 & 31.9\\
         & Temporal pooling & ResNet-50 & 32 & 33.8\\
         & TRG & ResNet-50 & 16 & 35.8 \\
         & TRG & ResNet-50 & 32 & \textbf{38.4} \\
         \bottomrule
        \end{tabular}
        \end{center}
        \vspace{-3mm}
        \label{tab:abla-sampling}
\end{table}

    \noindent\textbf{Effect of Different Backbone.}
    To measure the importance of context interaction of the temporal features in different backbones, we can plug our temporal reasoning graph into existing 2D CNNs and 3D CNNs architectures and inspect its activity recognition performance. For this, three temporal reasoning graphs were injected into the existing backbone at the top with almost the same setting. For fair comparison with existing methods, the 2D architectures, BnInception \cite{ioffe2015batch} and InceptionV3 \cite{szegedy2016rethinking}, are stemmed from TSN \cite{wang2016temporal}; and the 3D architectures, ResNet-50 \cite{he2016deep} and ResNet-101, are processed the same with non-local \cite{wang2018non} using I3D \cite{carreira2017quo} technique, which inflate the 2D convolution weights into 3D convolution. All the results are listed in Table \ref{tab:compare_smth}.  
    
    \noindent\textbf{Model Complexity.}
    Our approach has multi-head temporal graphs for temporal reasoning and a temporal semantic aggregator for features fusion, and the model can be deployed to the existing backbone. It has perfect scalability and provides a balance between performance and complexity. To figure out the time complexity of our model, we report the floating point operations per seconds (FLOPs) in a single clip to demonstrate the cost (shown in Table \ref{tab:compare_smth}). The additional cost and parameters induced by our module are mainly contained in the convolution kernels of the temporal similarity calculation and graph convolution parameters, followed by temporal semantic relation aggregation. More precisely, the parameters introduced by our temporal reasoning graph is $\sum_{l}\left(N_{l} C_{l}^{2}+9 N_{l} C_{l}^{2}+N_{l}^{2}\right)$, where $l$ denotes which stage we plug our block into the backbone, $C_{l}$ represents the dimension of the output channels, and $N_{l}$ denotes the numbers of the graph in this stage. The first term of the equation calculates the parameters in temporal similarity measure, the second term of the equation calculates the parameters in the spatial transformation of graph convolution, and the last term of the equation calculates the parameters in the temporal relation aggregator.
    
\subsection{Visualization}

    \begin{figure*}[h]
    \centering
        \includegraphics[width=0.75\textwidth]{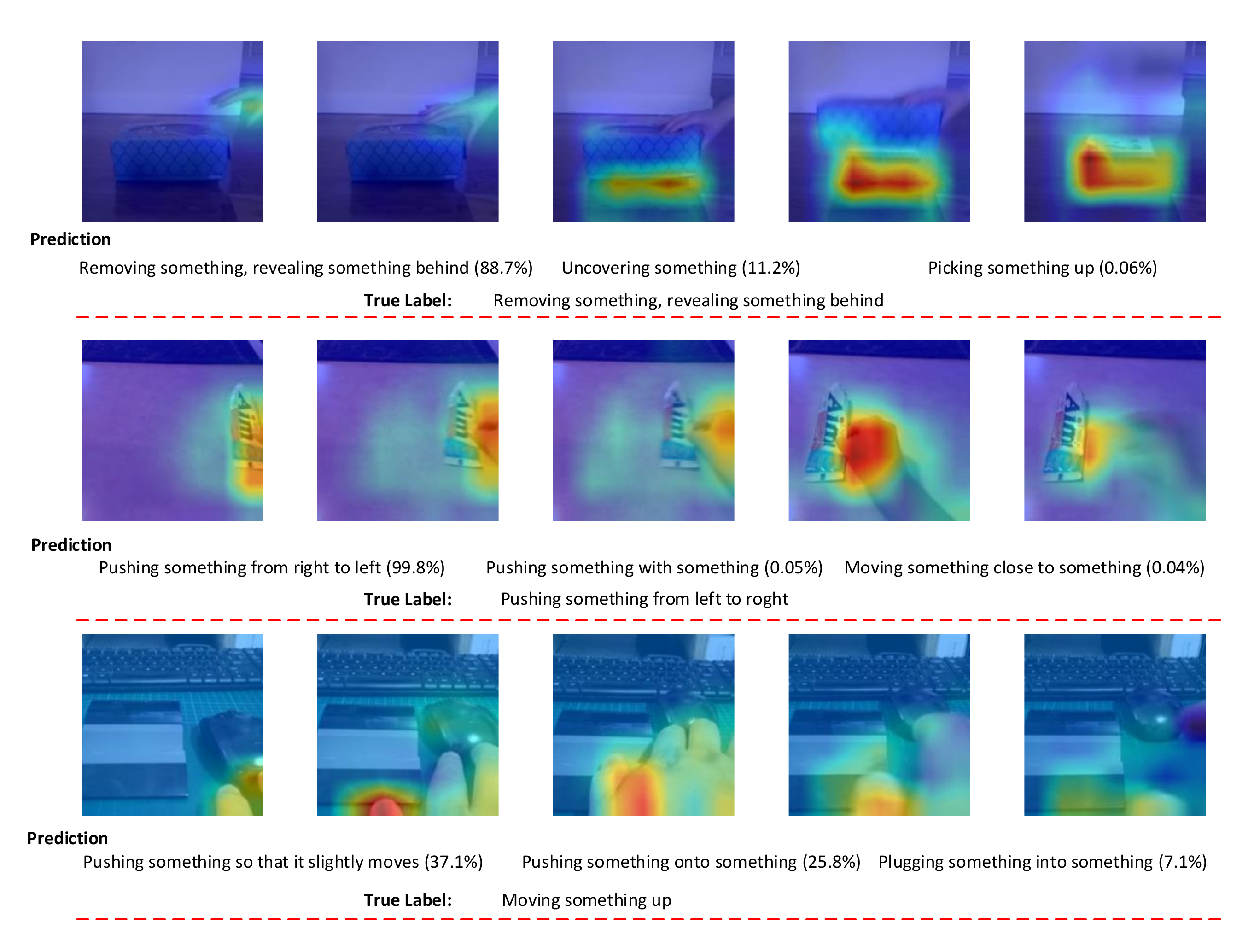}
        \caption{Visualization of ``CAM'' \cite{zhou2016learning} generated by our model. The maps highlight the discriminative region for action classification. The color in red denotes high importance for recognition. We also list the top-3 prediction scores of those examples.}
        \label{fig:cam}
    \end{figure*}
    %\vspace{-1cm}
    
    %\vspace{-0.9mm}
    \begin{figure*}[h]
        \centering
        \subfigure[Average pooling]{
        \begin{minipage}[t]{0.31\textwidth}
            \includegraphics[width=5.9cm]{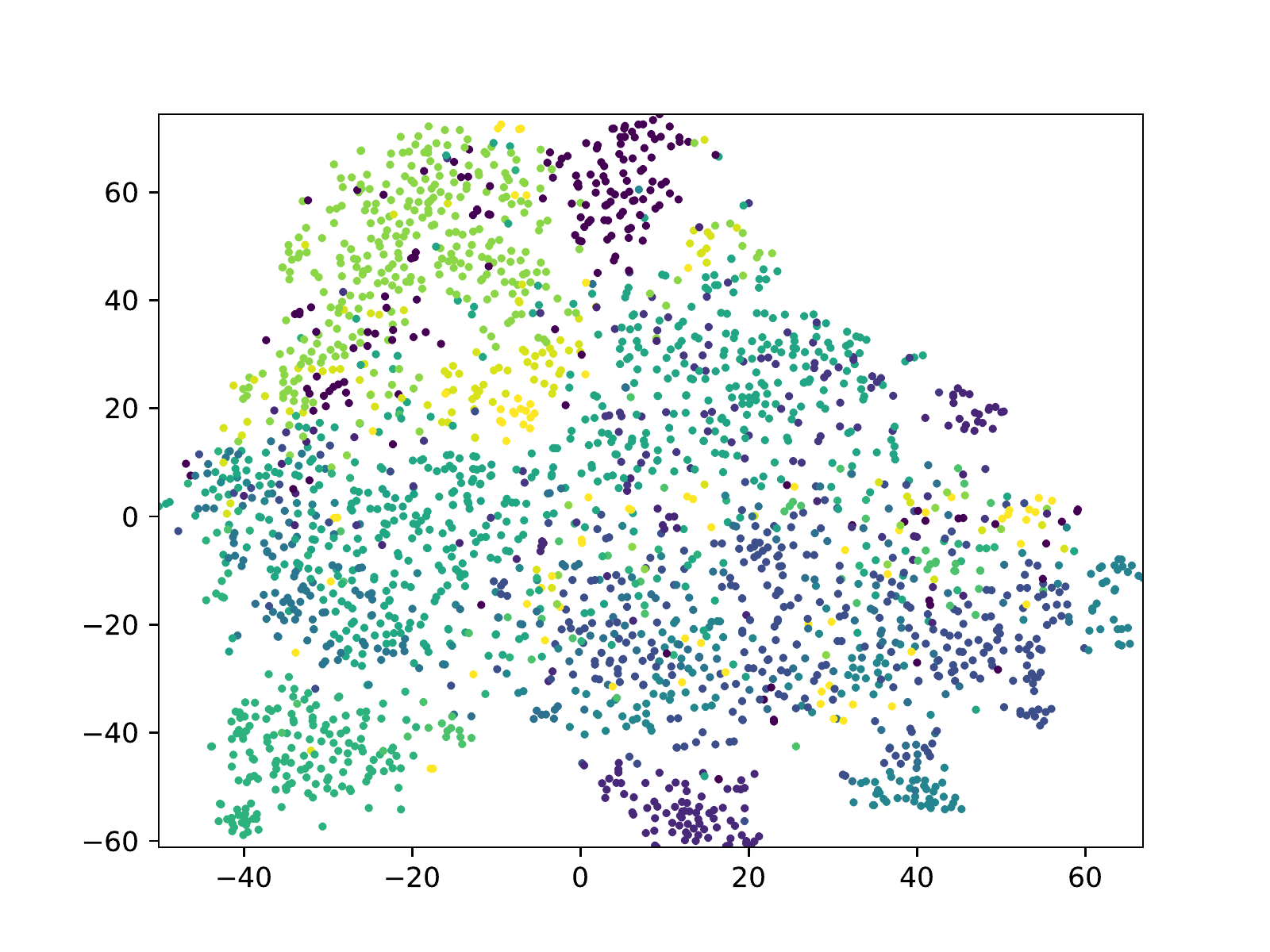}
        \end{minipage}}
        \subfigure[Three head adjacent matrix]{
        \begin{minipage}[t]{0.31\textwidth}
            \includegraphics[width=5.9cm]{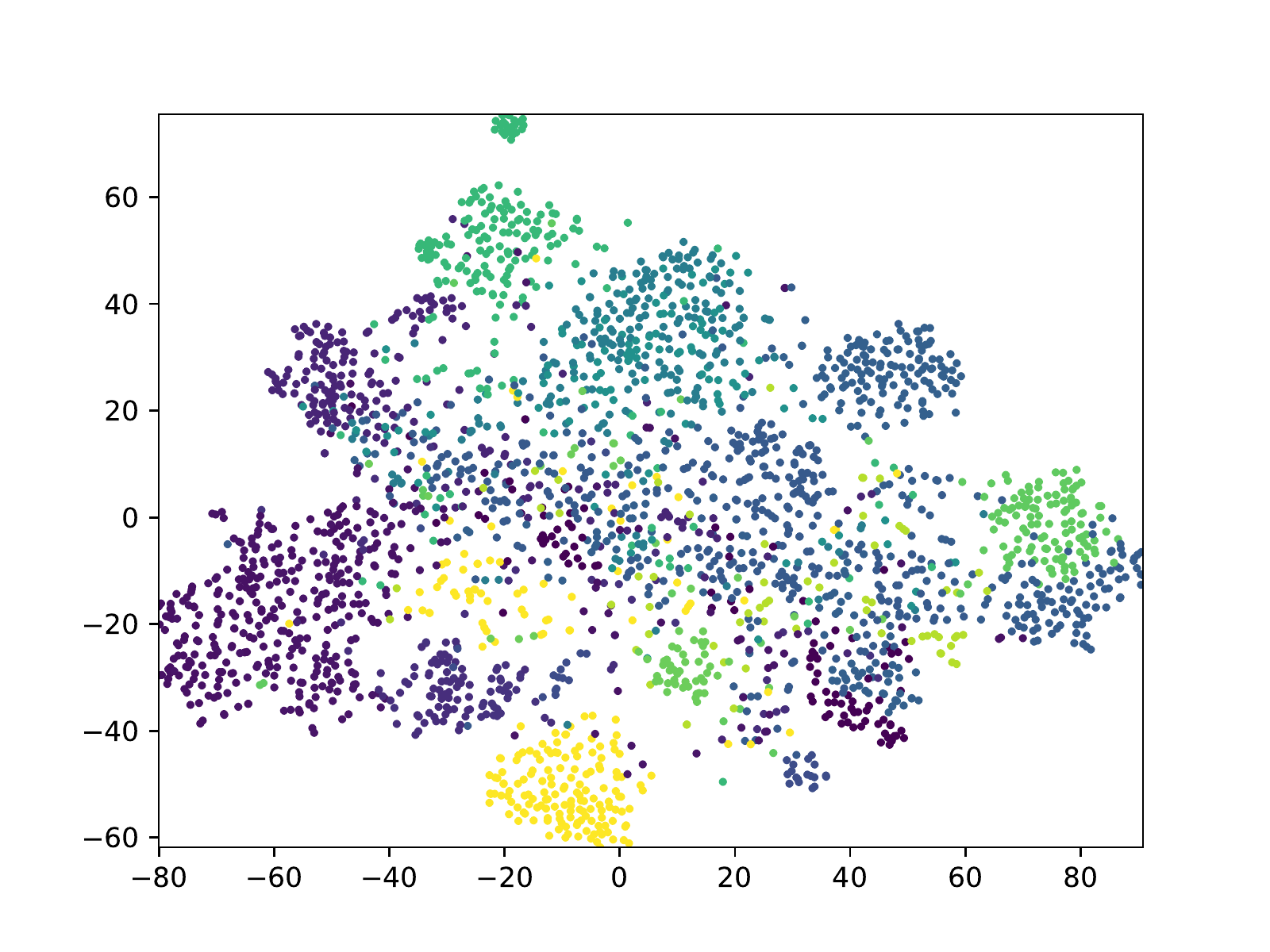}
        \end{minipage}}
        \subfigure[Six head adjacent matrix]{
        \begin{minipage}[t]{0.31\textwidth}
            \includegraphics[width=5.9cm]{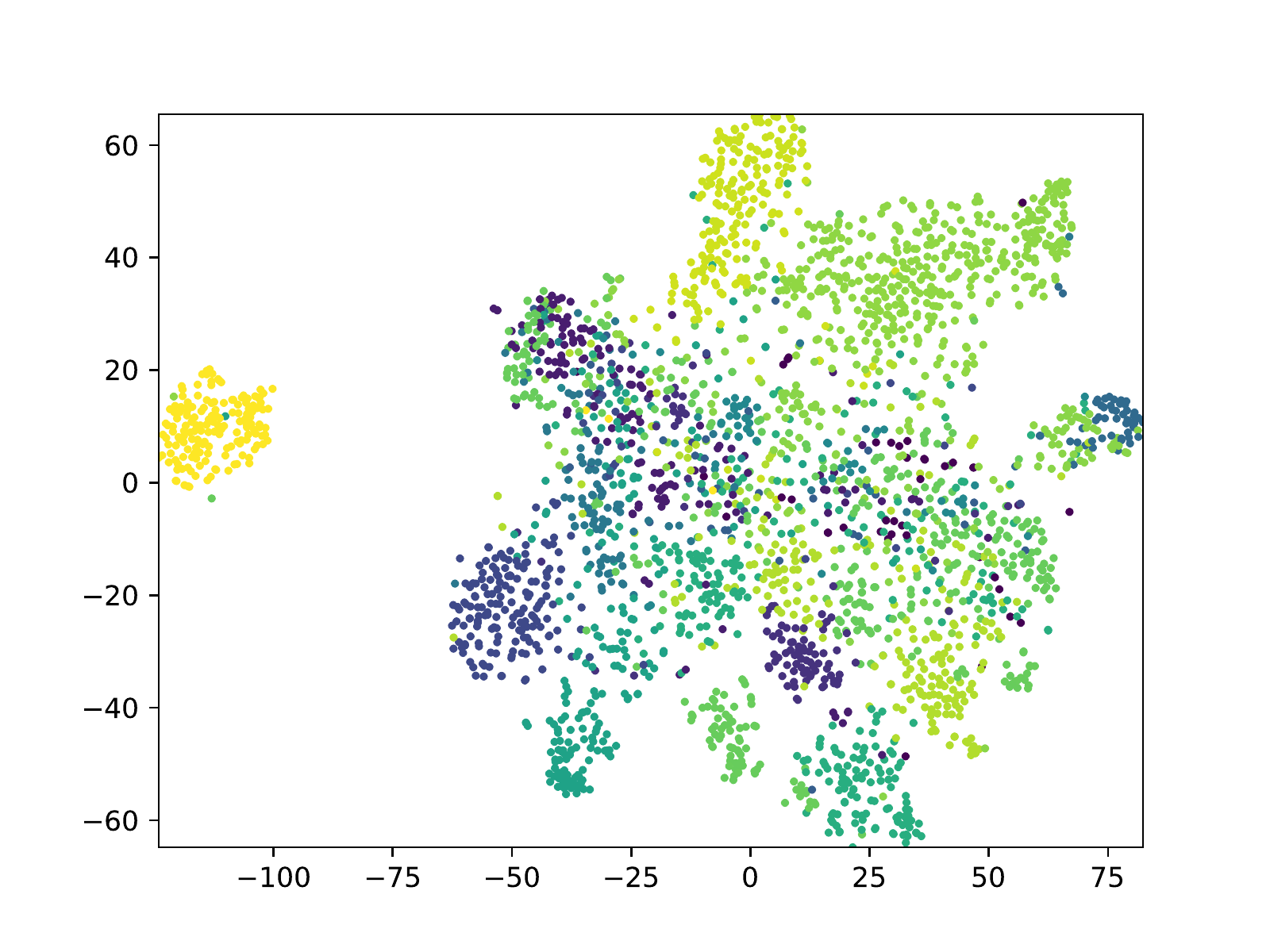}
        \end{minipage}}
        \caption{t-SNE visualization of the video representation on Something-Something V2 validation set. The representations are extracted based on Inception architecture with different settings: average pooling temporal features, 3 head adjacent matrix, and 6 head adjacent matrix. Different colors denote different action class and each point represents an embedding of a video.}
        \label{fig:tsne}
    \end{figure*}
    
    \noindent\textbf{CAM Visualization.} To get further insights into what our network learns, we visualize the CAM \cite{zhou2016learning} on Something-Something V2 dataset. CAM can visualize the most discriminative parts which will highlight the interesting region when classifying an activity of a clip. We randomly sample three video clips from the validation set and evaluate them by the trained model. To understand the primitives our model used for representing actions and visualize interesting class information, we show the output of CAM on the select samples. The results are shown in Figure \ref{fig:cam}. The highlight regions that correspond to the receptive field give us some insight into what the model cares about and indicating that spatio-temporal features are learned effectively.
    
    In Figure \ref{fig:cam}, we also list the top-3 prediction probability scores of the three examples. We observe that the model can perfectly recognize the first two cases, but it fails in the last example. For this kind of fine-grained action, the model still has trouble in classification. The score of this case, shown in the figure, predicts almost equally. However, from the prediction, we can not definitely declare that the result is wrong.

    \noindent\textbf{Visualization of The Learned Representation Distribution.} To visualize the distribution of the learned features, we apply the t-SNE tool (Principle Component Analysis for dimension reduction with 5000 iterations) for embedding the sequence representation extracted from our model with a different adjacent head number. The experiment was conducted on Something-Something V2 dataset with 20 randomly selected class which around 3K samples. The distribution difference between Figure \ref{fig:tsne} (a) and (b), (c) reveals that the model can learn more discriminative embedding with the temporal reasoning graph. Moreover, as shown in Figure \ref{fig:tsne} (b) and (c), through the multi-head adjacent matrix, the proposed model could separate sample points into several semantically discriminative clusters.
    
\section{Conclusion and Future Work}
\label{sec:conclusion}
    In this work, we presented a novel temporal reasoning graph module that captures the temporal relation of a video sequence to address the long-range temporal dependencies for activity recognition. Meanwhile, we proposed the multi-head adjacent matrix to investigate the multi-kinds relations of a video sequence. Additionally, a multi-head temporal relation aggregator was designed to automatically learn the importance of different sequence state in different graphs for comprehensive video-level feature learning. Benefiting from these two novel module designing, the proposed model can capture multi-kinds temporal relation with different scale and temporal span. We evaluated the proposed model on Something-Something and Charades datasets and established competitive results compared with existing methods.
    
    We hope the proposed temporal reasoning graph module will boost performance on various video understanding tasks. Furthermore, we plan to investigate the power of GCNs in video context correspondence learning with a self-supervised manner.

\section*{Acknowledgment}
This work was supported in part by the National Natural Science Foundation of China under grants No. 61502081, 61602089, 61632007 and the Sichuan Science and Technology Program 2018GZDZX0032, 2019ZDZX0008 and 2019YFG0003.


% \ifCLASSOPTIONcompsoc
%   % The Computer Society usually uses the plural form
%   \section*{Acknowledgments}
% \else
%   % regular IEEE prefers the singular form
% \fi




%\bibliographystyle{IEEEtran}
\bibliographystyle{ieee}
\bibliography{TRG.bib}

\end{document}